\UseRawInputEncoding
\documentclass{llncs}
\usepackage{makeidx}  
\usepackage{mathtools}
\usepackage{algorithm}
\usepackage{multirow}
\usepackage{tabularx}
\usepackage{etoolbox} 
\usepackage{graphicx}
\usepackage{cite}
\usepackage{caption}
\usepackage{subcaption}
\usepackage{setspace}
\makeatletter
\patchcmd{\ps@headings}{\rlap{\thepage}}{}{}{}
\patchcmd{\ps@headings}{\llap{\thepage}}{}{}{}
\makeatother
\usepackage[a4paper,margin=1.5in,footskip=0.3in]{geometry}

\begin{document}
%
%

%
\mainmatter              
\title{Robot to Human Object Handover using Vision and Joint Torque
Sensor Modalities}
\titlerunning{Hamiltonian Mechanics}  
%
\author{Mohammadhadi Mohandes\inst{1} \and Behnam Moradi\inst{2}
Kamal Gupta\inst{1} \and Mehran Mehrandezh\inst{2}}
\authorrunning{Ivar Ekeland et al.} 
%
\tocauthor{Mohammadhadi Mohandes, Behnam Moradi, Kamal Gupta, Mehran Mehrandezh}
\institute{School of Engineering Science, Simon Fraser University, Canada,\\
\email{{mmohande,Kamal}@sfu.edu}
\and
Faculty of Engineering and Applied science, University of Regina, Canada,\\
\email{bmn891,mehran.mehrandezh@uregina.ca}}

\maketitle              
\vspace{-0.5cm}
\begin{abstract}
We present a robot-to-human object handover algorithm and implement it on a 7-DOF arm equipped with a 3-finger mechanical hand. The system performs a fully autonomous and robust object handover to a human receiver in real-time. Our algorithm relies on two complementary sensor modalities: joint torque sensors on the arm and an eye-in-hand RGB-D camera for sensor feedback. Our approach is entirely implicit, i.e., there is no explicit communication between the robot and the human receiver. Information obtained via the aforementioned sensor modalities are used as inputs to their related deep neural networks. While the torque sensor network detects the human receiver's ``intention'' such as: pull, hold, or bump, the vision sensor network detects if the receiver's fingers have wrapped around the object. Networks' outputs are then fused, based on which a decision is made to either release the object or not. Despite substantive challenges in sensor feedback synchronization, object and human hand detection, our system achieves robust robot-to-human handover with 98\% accuracy in our preliminary real experiments using human receivers.
 
\vspace{-0.4cm}
\end{abstract}

\vspace{-0.5cm}
\section{Introduction}
Human-Robot Interaction (HRI) is a wide and diverse area of research. A robot-to-human (R2H) handover task, as a sub-topic of HRI, is defined as a mission of transferring an object from a giver (an autonomous robotic system) to a receiver (a human operator). Our focus is on direct R2H tasks where the robot directly delivers the object in human receiver's hand. A successful object handover mission happens when the giver makes sure that the receiver has fully taken possession of the object and feels safe to let go of the object. Failure in R2H object handover often occurs when there is a wrong interpretation (by the giver) of the actions applied on the object by the receiver. Hence, the problem of failure detection plays an important role in a successful handover and requires detecting human ``intention'' accurately \cite{ortenzi2021object}. Safety, reliability, and robustness of the object handover therefore depend directly on the sensor modalities  such as vision, force/torque, tactile, and their respective interpretation. In particular,  the physical contact phase between the receiver and the object poses a serious challenge. Early releasing is considered to be a safety challenge while late releasing can cause higher interaction forces \cite{chan2012grip}. 
In this paper, we use two key sensor modalities, joint torque sensors of the arm and an eye-in-hand RGB-D camera for accurate determination of when the robotic hand should release the object, hence resulting in a robust R2H handover. More specifically, the joint torque time series data is used to train a CNN network that predicts the receiver's action/intention (pull, pull-up, push, bump, hold, and no action) during the contact phase, and a second network, a Single Shot multibox Detector (SSD) \cite{liu2016ssd} to detect fingertips and objects in real-time in order to robustly determine the physical contact between the human receiver's hand and the object. The outputs of the two CNNs are then fed to a finite state machine, that in essence, results in a release command only if the vision pipeline detects contact between the receiver's fingers and the object, and the torque pipeline detects a pull, pull-up, and hold. Our initial experimental results with human receivers are extremely positive showing a 98\% success rate in R2H tasks.
While joint force/torque sensors have been used in previous works on R2H handover tasks [\cite{sileo2021vision} and \cite{eguiluz2017reliable}], and have been combined with a specialized  simple optical sensor, designed specifically to detect object motion [\cite{parastegari2016fail} and \cite{parastegari2018failure}]\footnote{ In fact, that work used a specialized simple optical sensor precisely because they mention the unacceptable amount of computation time that would be taken for processing RGB images, a problem that we solve via the use of SSD network.}, the key contributions of our work are: i) we use joint torque sensors' data in a novel way, i.e., we use a time series of joint torques to detect human receiver's action/intention for R2H handover tasks, (ii) we use an eye-in-hand RGB-D camera and detect finger contacts with the object in real-time (30fps), and iii) to combine i) and ii) as an algorithmic fusion approach to make a robust RELEASE decision. Our preliminary real experiments with human receivers show a 98\% success rate. We also compare our method's success rate with some existing R2H systems \cite{shi2013model,choi2009hand, grigore2013joint, koene2014experimental, prada2014implementation} that have used success rate as an evaluation metric. Please note that some other works report human satisfaction surveys to evaluate  R2H systems, e.g., \cite{parastegari2018failure,chan2013human}, which is different than the success rate metric that we report.
 
 The rest of this paper is organized as follows:
Section II presents a comprehensive literature review on vision-based and force/torque-based object handover. Section III presents the methodology and the algorithmic foundations of our work. Section IV shows the experimental results. Finally, section V presents conclusions and future works.

\vspace{-0.4cm}
\section{Related Work}
\vspace{-0.3cm}
 A key challenge in robot-to-human object handover (R2H), unlike robot-to-robot handover (R2R), as we mentioned earlier in the introduction is that there is no real-time sensor data exchanging between the human receiver and the robot other than onboard sensors of the robot. The robot (in our case, a 7-DoF Gen 3 Kinova arm) with a 3-fingered mechanical hand (Schunk SDH) has  two sensor modalities: i) joint torque sensors and ii) an eye-in-hand RGB-D camera. Therefore, we have focused on these two modalities in our current work and our literature review below focuses on R2H handover works that use one or both of these two modalities to understand the intention of the human receiver in R2H handover tasks.  The research community has attacked this challenge using two main approaches: Vision-based and force/torque based. We first outline some general vision-based approaches from the machine vision community and then present the R2H literature.

\subsection{Vision-based computations in general}\label{sec:vision-general}

In the machine vision community, vision data has been used in a variety of ways - detecting human gaze, human body configuration,  human hand, and object detection. A key requirement in R2H tasks is to accomplish this in real-time. Detecting human's hand and the object in real-time is investigated in \cite{hasson2019learning}, \cite{zimmermann2019freihand}, and \cite{hampali2020honnotate}. Human body tracking and its related pose with respect to the object is investigated in \cite{ijspeert2002movement} and \cite{schaal2005learning}.

From the perception perspective, Single Shot multibox Detector (SSD), a CNN-based network architecture that was introduced by Lio et al. \cite{liu2016ssd} is particularly appealing for object detection for R2H tasks and we adapt it for our application along with a bounding box regression algorithm has been taken from Google’s Inception Network. The SSD network combined with bounding box regression is able to outperform Faster R-CNN (another competing neural network-based architecture for object detection) in accuracy and in speed to obtain 59+ fps. SSD is capable of detecting multiple objects. Unlike R-CNN methods, it propagates the feature map in one forward pass throughout the network. This is the main reason that SSD is able to operate in real-time and handle object overlap in the data points. SSD uses a pre-trained network as a basic net which is trained on the ImageNet dataset. There are multiple convolutional layers and each one of them is individually and directly connected to the fully connected layer. A combination of SSD and bounding box regression allows our network to detect multiple fingertips with different scales \cite{liu2016ssd}.

\vspace{-0.3cm}
\subsection{Vision-Based Object Handover}
\vspace{-0.1cm}
Real-time object detection and pose estimation is a challenging problem associated with vision-based object handover \cite{yang2020human}. To address this problem, minimum jerk trajectory algorithm is used to predict the receiver's intention \cite{li2015predicting}. Gaussian process is also used to estimate human motion in object handover scenarios. Gaussian mixture regressor is proposed by Lue et al. to predict human body motion while trying to receive the object from the robot \cite{luo2018unsupervised}. The receiver's hand pose estimation can also be used to estimate the approaching speed of the hand \cite{controzzi2018humans}.

Vision-based approaches fundamentally utilize classical computer vision and more recently, deep-learning techniques to detect the object using an RGB-D camera, as mentioned above in Section \ref{sec:vision-general}. The output of vision-based techniques is mainly detected bounding boxes around the object in RGB image and estimated pose of the object in the point cloud. 

Strabala et al. \cite{strabala2013toward} reported a comprehensive investigation of robot-to-robot and robot-to-human object handover scenarios. The first scenario is to understand the way humans exchange physical objects followed by recording the physical behavior of both the giver and the receiver. The next scenario was to codify the human-to-human object handover in order to implement it as a robot-to-human object handover algorithm. In this scenario, when the robot is the giver and the human operator is the receiver, the robot should go through three crucial steps of detecting a human's body, eye gaze, and hands to confirm that the human operator is ready to receive the object. In our case, since the camera is wrist mounted (eye-in-hand), the fingertips and object are the natural choices to be detected.  

Grigore et al. \cite{grigore2013joint} focused on adding eye gaze detection and head orientation as a user intention model to an HMM-based R2H object handover. Their results demonstrate significant improvements in object handover success rate  by integrating this vision sensor-based feedback into the robot’s control system. The use of eye gaze has been also promoted in \cite{shi2013model} where R2H handover has been tested on robots distributing flyers to uncooperative passing pedestrians to make a better successful ratio. Koene et al. \cite{koene2014experimental} and Prada et al. \cite{prada2014implementation} developed a color segmentation method with a Kinect sensor to estimate the human's hand location during the handover process. \cite{choi2009hand} compared the success rate between  direct delivery and indirect delivery, implemented on an EL-E robot equipped with a force/torque sensor and a laser-pointer interface with a camera detecting the 3d location of handover.
 Sileo et al. \cite{sileo2021vision} developed a vision-based robot-to-robot object handover algorithm that was able to go through predetermined steps to deliver a known object from one robot to another without using explicit communication between two agents. Other than visual data, they also incorporated force/torque data from both end effectors to increase the robustness of the handover process. For object detection, Faster-RCNN and YOLO deep learning networks were utilized to detect objects in real-time. A key distinction in our work is that we utilized data from all joint-level torque sensors to classify the receiver's intention while performing the handover mission. In addition, it is conjectured that better results can be obtained by fusing vision and torque data in order to perform a robust and safe object handover mission. However, when it comes to object detection, sensor data synchronization poses a serious challenge, because image processing is normally slow. Our SSD fingertips detection performs at 59+ fps and is more accurate than YOLO and Faster-RCNN, thereby meeting the sensor synchronization challenge. 

\vspace{-0.4cm}
\subsection{Force/torque sensor based Object Handover}
In \cite{parastegari2016fail} a handover controller is proposed and implemented on a 2-finger gripper installed on Baxter robot. A key novelty of the system is that it  re-grasps a fast slipping object via a closed-loop feedback controller that detects the downward acceleration of the object via a specially designed simple optical sensor attached to the wrist of the gripper. The proposed controller is claimed to be smooth and effortless and able to release the object in a certain range of pulling directions. A key distinction of our work is that our torque-based detection scheme for receiver actions is far more comprehensive and robust, i.e., we incorporate six action classes (i.e. push, pull, pull-up, hold, bump, and no action) in a torque time series dataset which we use to train a CNN network for action classification in real-time. Moreover, our approach uses real-time vision with a standard off-the-shelf camera (rather than a specialized designed optical sensor) to detect the full grasp by the receiver before releasing the object.

 \cite{chan2013human} proposes a release controller deployed on a PR2 robot with the help of lessons learned from human-to-human object handover experiments.  The giver’s and receiver’s load force and grip force on the object are measured during handover. It is shown that the receiver’s grip force is proportional to the load force on the object and a linear release controller is developed that attempts to reduce the gripping force with respect to the load force. The handover in this controller was carried out only in the vertical direction.

Another proposed approach to detect the action is to use the sensory data to classify various events affecting the object. In our previous work, \cite{davari2019identifying}, we introduced a classification algorithm  to categorize the different actions  the receiver applies on the object, such as pull, push, bump, and hold during handover using tactile data. A Bag of Words (BOW) algorithm was used for feature extraction, k-means clustering was used for dimension reduction and a Support Vector Machine (SVM) was used for classification. In future works, we would like to incorporate tactile data as well.

 \cite{eguiluz2017reliable} explores cases where the receiver may make unwanted contact (manipulation or disturbance) with the object, such as unwanted rotation or pushing where the object should not be released and differentiates them from legitimate actions such as pulling where the object should be released by the giver. Disturbances are detected via a probabilistic model from tactile data and compensated for by an effort controller via appropriate finger movements to maintain a stable grasp.
\vspace{-0.5cm}
\section{Methodology}
\vspace{-0.3cm}
Our object handover approach uses the information obtained via two sensor modules on our Kinova 7-DOF arm, namely joint-level torque sensors and an RGB-D camera mounted on the robot's end effector. We use a CCN-based classifier to classify the torque data, while the RGB-D information is utilized by an SSD object detector to detect and estimate the position of human fingers approaching and/or grasping the object. The CNN-based classifier takes the joints' torque data and returns the output class (e.g. pull, push, bump, pull-up, hold, and no-action). The SSD object detector generates a bounding box for each detected finger along with their positional information with respect to the object. In this section, the structure of our dataset and network architectures are explained in detail followed by our fusion function, which is utilized to take the torque-based classification results along with that under the finger detection (bounding box and pose) for making a binary RELEASE decision for the object handover mission. Please note that although we have used the white bottle as a canonical object (bottle image is shown in Fig. \ref{camera-actions}b), our approach is not specific to a given object and generalizes to a wide class of objects. The torque-based CNN classifier uses the joint torques (and not explicit object geometry) and we have trained our SSD model for different objects in hand with great success. Multi-object experiments will be carried out in the next phase of the project. 
\vspace{-0.4cm}
\subsection{Dataset and Annotation}
\vspace{-0.2cm}
In order to provide sufficient data points for the training purpose, we went through a comprehensive data collection and annotation process, which is explained below.
\vspace{-0.4cm}
\subsubsection{Torque Dataset}

\begin{figure}
    \centering
    \includegraphics[width=\textwidth]{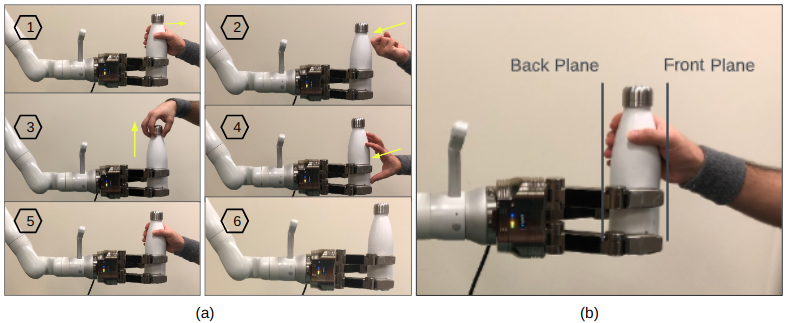}
    \caption{(a)Actions used in collecting dataset. 1: Pull, 2: Bump, 3: Pull-up, 4:
Push, 5: Hold, 6: No Actions and (b) Kinova RGB-D camera is used to detect fingers lying in between the front and back planes of the object.}
    \label{camera-actions}
\end{figure}

We used a Kinova Gen3 7-DOF robotic arm that consists of a stationary base, 7 revolute joints equipped with torque sensors, and a vision sensor mounted on the robot's end-effector. Joint-level torque data are used for torque-based action classification. Each torque sensor provides times-series data sampled at 40 Hz. The torque data are reported by Kinova's API and its ROS driver in a position-control mode. Fig. \ref{camera-actions}a illustrates the actions we used to collect data. We have asked 10 volunteers (male: 5, female: 5) to participate in our data acquisition process carrying out 6 actions. Each action was repeated 30 times by each volunteer.

\subsubsection{RGB Dataset}
Our vision dataset consists of a comprehensive set of RGB images that are collected by an eye-in-hand RGB-D camera. Fig. \ref{camera-actions}b illustrates the camera's setup. In order to maintain the data diversity and to increase the number of data points, we asked 10 volunteers (male: 5, female: 5) to participate in the robot-to-human object handover tasks, and recorded their actions during the experiments. Afterwards, we used the labelImg Python package to annotate the RGB images based on the VOC standard. Our target object to identify and classify would be the fingertips of the human receiver.

\vspace{-0.4cm}
\subsection{Vision-based Object Handover}
\vspace{-0.2cm}
Detecting the human’s intention based on the information obtained via our RGB-D camera is challenging. Real-time inference process and accurate detection are essential for performing a reliable and precise object handover. Our vision algorithm is based on a Single Shot Multibox Detector (SSD), which is able to operate in real-time and detect multiple objects, despite partial camera occlusion. Accordingly, we implemented a complete pipeline, which can be trained on any custom dataset for real-time object detection. Our SSD-based training pipeline is made open source and available to the research community. (\cite{Behnam_ssd_train2021} and \cite{Behnam_ssd_ros2021}).

\begin{figure}
    \centering
    \includegraphics[width=\textwidth]{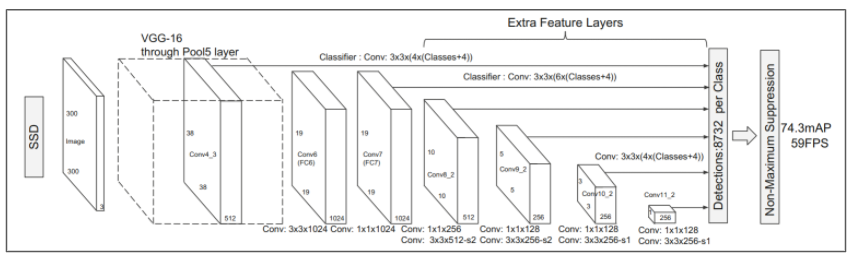}
    \caption{SSD Network Architecture. \cite{liu2016ssd}}
    \label{ssd-network}
\end{figure}

Single Shot Multibox Detector (SSD) is a deep neural network specifically designed for multiple object detection from RGB images. It is designed based on a feed-forward convolutional network that is able to utilize multiple-scale feature maps for object detection. The base network is associated with convolutional feature layers to allow the network to detect objects of multiple scales. Fig. \ref{ssd-network} illustrates the SSD network architecture.

The SSD model adds feature layers to the end of a base network, that further enables it to predict the offset among bounding boxes, aspect ratios, and their confidence scores. A set of convolutional filters are placed at each feature layer to detect a fixed number of bounding boxes in the image. SSD is different from other object detectors in the way that the ground truth information is assigned to the predetermined outputs. 
From the training perspective, SSD is specifically designed based on  multibox objective \cite{szegedy2014scalable}, which enables it to detect objects of multiple categories. The total objective-loss will include two parts, namely a localization (loc) and a confidence loss (conf) where $x_{ij} ^{p}$ indicates binary matching criteria to match i-th detected box to the j-th ground truth box of category p. Also, n the matching strategy above, we can have $\sum_{n=1} x_{ij} ^{p} >=1 $.
\begin{equation}
    L(x,c,l,g)= \textstyle\frac{1}{N} (L_{conf}(x,c) + \alpha L_{loc}(x,l,g))
\end{equation}

\vspace{-0.2cm}
The number of default boxes is indicated by N. “l” denotes the predicted bounding box and “g” is the ground truth parameter. A combination of these two is used to calculate the loss value. To calculate the confidence loss, a combination of softmax-loss over the classes confidence “c” and the weight “alpha”. 
In our training pipeline, we performed several experiments that are summarized in Table I. Through the training experiments, we realized that data augmentation resulted in a lower loss value and a higher mAP. Therefore, we picked this model as the accurate one to perform our vision-based object handover.

\vspace{-0.3cm}
\subsection{Torque-based object handover}
\begin{figure}
    \centering
    \includegraphics[width=\textwidth]{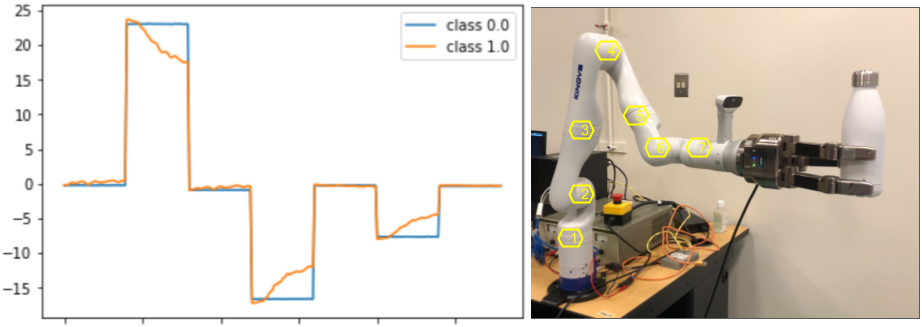}
    \caption{Time series of torque data - All sensors are included in the time series which is converted to a 1D vector to be used as an input to the CNN network.}
    \label{torque-samples}
\end{figure}

The Kinova 7-DOF arm comes with joint-level torque sensors. From the software perspective, Kinova's ROS Kortex package publishes each sensor data on a specific topic. Therefore, we have access to each sensor data in real-time which is then converted to a time series input data. A sample of our torque time series is illustrated in Fig. \ref{torque-samples}. Afterwards, this time series is converted into a 1D vector by concatenating all seven joint-level sensor data  and is used as input data to our CNN classifier. 

\begin{figure}
    \centering
    \includegraphics[width= 0.6\textwidth]{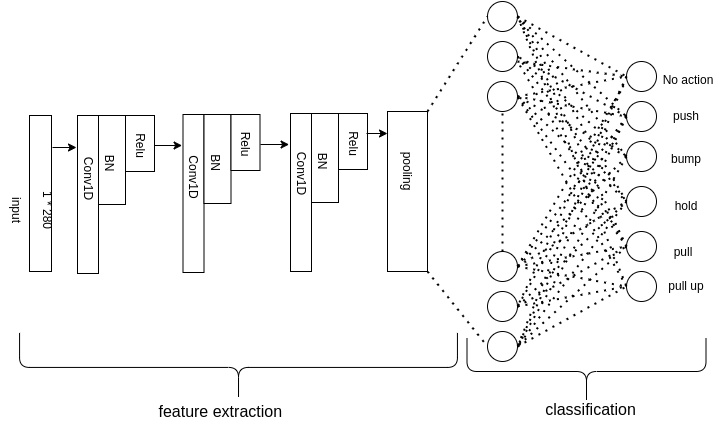}
    \caption{CNN network architecture - torque data classification}
    \label{cnn-arch}
\end{figure}

As shown in Fig. \ref{cnn-arch}, the network consists of 4 blocks. The first 3 blocks have the same configuration. The output of a convolutional layer is fed into a batch normalization layer which is followed by a Relu activation at the back end of the model. Since torque data are converted to a 1D vector, the kernel size needs to be a 1-tuple specifying the length of the convolution window. We choose {3} as our kernel size and the number of filters specifying the number of kernels to convolve with the input volume as {64}. Each kernel operation will provide a 1D activation map.  
Depending on the robot's configuration and the actions/events, torque readings can change from 0 to 30 N.m. A batch normalization was applied in order to generalize the torque data and facilitate faster convergence. 
The whole complete network is made by concatenating 3 blocks with a global average pooling layer at the end followed by a softmax classifier to label the output. 
We collected a dataset of 6 actions, i.e., bump, pull, pull-up, pull down, push, and no action. Each action includes 300 training data sampled for one second at 40 Hz.

\vspace{-0.4cm}
\subsection{Fusion algorithm}

\begin{figure}[h!]
  \centering
  \includegraphics[scale=0.3]{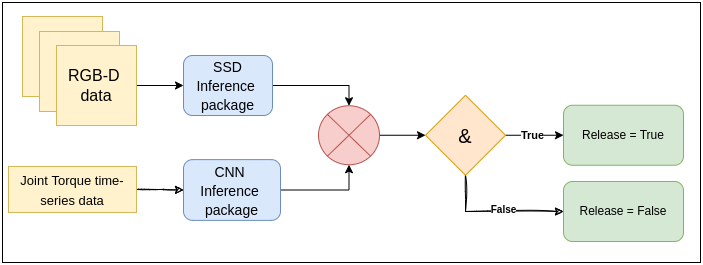}
  \caption{Algorithmic fusion of torque and vision data in our object handover mission.}
  \label{fusion-diagram}
\end{figure}

The fusion algorithm takes the output of the torque-based CNN classifier and the output of the SSD fingertips detector in real-time. From a programming perspective, it is written as a ROS package that is named "torque-vision-function" package. It subscribes to the topics of the torque-based CNN classifier and SSD fingertips detector nodes. In order for the torque-vision-function node to operate in real-time, the topic-callback-synchronization is also implemented by our fusion ROS node. Figure \ref{fusion-diagram} illustrates the fusion algorithm. Utilizing the point cloud data, the SSD fingertips detector also estimates each fingertip's position with respect to the camera frame. For the SSD fingertips detector to output the release signal, it is necessary to detect at least three fingers, including the thumb, and the fingertips' positions need to fall between the front and back planes of the object. Fig. \ref{camera-actions}b illustrates a specific object and its planes. The torque-based CNN classifier is utilized to detect the contact between the hand and the object as well as the direction of the applied force. A simple "AND" function is used over the outputs of the SSD fingertip detector and the torque-based CNN classifier to determine if the gripper should release or not. Further details are given in Section \ref{sec:fusion-function}.

\section{Experiments and Results}
Table I presents the success and failure criteria that we used to determine whether an object handover mission was successful or not. In this section, we report on our experimental results under 3 scenarios of object handover: 1- torque-based, 2- vision-based, and 3- fusion-function based. The overall results of all experimental scenarios are reported in Table II. In each experiment scenario, we used 30 volunteers to perform the actions defined in Fig. \ref{camera-actions}a.

\vspace{1cm}
\begin{table}
\centering
\caption{Each action in the object handover mission is evaluated using pre-defined success criteria.}
\label{table:1}
\begin{tabular}{ |c |c| } 
\hline \hline 
actions & success criteria \\
\hline
no action & Do not release the object \\ 
bump & Do not release the object \\ 
push & Do not release the object \\ 
hold &  Release the object smoothly \\ 
pull &  Release the object smoothly \\ 
pull-up & Release the object smoothly \\ 
\hline

\end{tabular}
\end{table}

\vspace{-1.5 cm}
\begin{table}
\centering
\caption{Our experiments are carried out based on three R2H object handover scenarios: 1- vision-based, 2- torque-based, and 3- fusion-based. "s" and "f" indicate "success" and "failure" respectively. We asked 30 volunteers to do six actions on the robot. The results are recorded based on the success criterion mentioned in Table I.}
\label{table:2}
\begin{tabular}{ |c|c|c|c|c|c|c|c| } 
\hline \hline 
actions  &\multirow{2}{4em}{number of trials} & \multicolumn{2}{|c|}{torque-based} & \multicolumn{2}{|c|}{vision-based} &\multicolumn{2}{|c|}{fusion-based} \\

&&s&f&s&f&s&f \\
\hline
no action & 30 & 30 & 0 & 30 & 0 & 30 & 0 \\ 
bump & 30 & 25 & 5 & 28 & 2  & 30 & 0 \\ 
push & 30 & 27 & 3 & 0 & 30 & 29 & 1 \\ 
hold & 30 & 24 & 6 & 28 & 2 & 29 & 1 \\ 
pull & 30 & 28 & 2 & 30 & 0 & 30 & 0\\ 
pull-up & 30 & 28 & 2 & 26 & 4 & 29 & 1 \\ 
\hline

\end{tabular}
\end{table}

\vspace{-0.5 cm}
\subsection{Torque-based object handover}
\subsubsection{Training pipeline}
In order to train a CNN classifier based on torque data, we collected a diverse dataset that consists of six pre-defined actions. Fig. \ref{camera-actions}a illustrates the actions used in our torque dataset. Human subjects were provided with safe instructions on how to perform each desired action in an object handover mission. Our ROS user interface provides recording instructions and the human subject starts implementing the action after seeing a prompting message on the screen. Afterwards, the recorded dataset was post-processed, in which we collect a one-second time window out of each individual run for training the CNN network.

\begin{figure}[h!]
  \centering
  \includegraphics[scale=0.3]{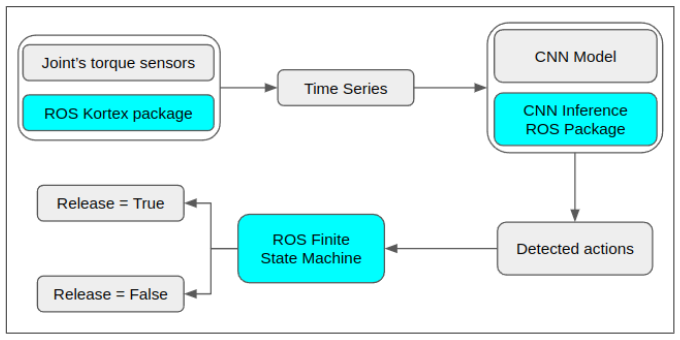}
  \caption{Torque-Based object handover. The Finite State Machine node commands the robot to release the object when a receiving action is detected from the torque sensors.}
  \label{CNN-Diagram}
\end{figure}

\subsubsection{Experiments}
Fig. \ref{CNN-Diagram} shows the block diagram of torque-based object handover algorithm. During the handover mission, the ROS node runs at 40 Hz and it relays one second of torque data to the CNN classifier. Once the inference process is finished, the output will be the RELEASE decision. According to our overall results reported in Table II, out of 180 missions (i.e., the total number of experiments), our success rate in torque-based object handover was 90\%. 

\subsection{Vision-based object handover}
\subsubsection{Training Pipeline}
Using our SSD training pipeline, we performed several training experiments. Through our training experiments, we realized data augmentation resulted in a lower loss value and higher mAP@0.5,0.7 in our large dataset.  Therefore, we picked the model that is trained using our augmented dataset which yields higher mAP@0.5 and lower overall training loss value.

\begin{figure}[h!]
  \centering
  \includegraphics[scale=0.42]{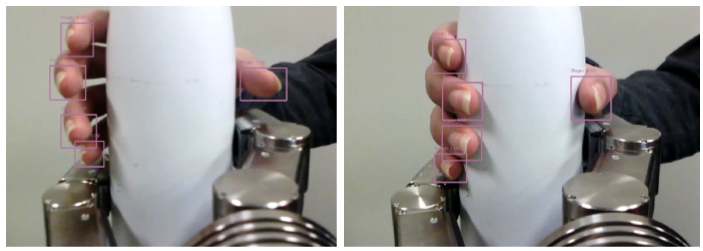}
  \caption{Real-time fingertips detection while the human receiver is trying to receive the object. Top Image: The object is not held by the human receiver. Bottom Image: The Object is held by the human receiver.}
  \label{inference_image}
\end{figure}

\subsubsection{Experiments}
\begin{figure}[h!]
  \centering
  \includegraphics[scale=0.35]{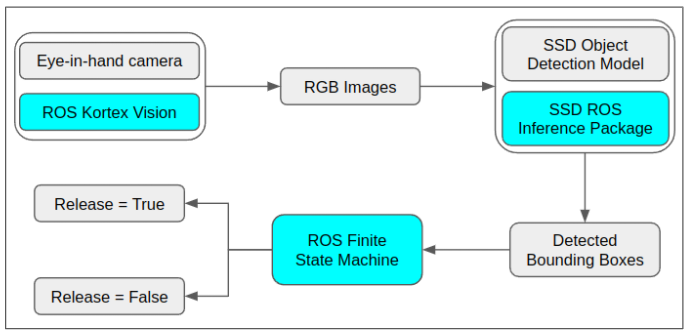}
  \caption{Block diagram of vision-based object handover. The Finite State Machine node allows the robot to release the object when all the fingertips are detected on the object.}
  \label{vision_diagram}
\end{figure}

Fig. \ref{vision_diagram} illustrates the block diagram of vision-based object handover algorithm. From the implementation perspective, our algorithm was designed to be compatible with the ROS platform. Therefore, every sub-algorithm was written in a specific ROS package. RGB images were  obtained from the camera using the ROS-Kortex-Vision package developed by KinovaRobotics. Afterwards, the RGB images (300x300 pxl) were published on a specific topic in real-time which was subscribed by our SSD-ROS-Inference package. This package uses the trained SSD model to establish an inference session and to publish the detected bounding boxes in 59+ fps. 
Using a 300x300 pixel resolution in images, the reported mAP@0.5 was 96\%. Also, our SSD model was able to run at 59+ fps, thus, suitable for real-time detection of fingertips. Our vision-based object-handover experiments were carried out by the same number of volunteers (i.e., 30). Fig. \ref{inference_image} illustrates a representative of those experiments. Once all the fingertips were detected, the FSM outputs a RELEASE decision that will be followed by the SDH gripper releasing the object. Out of 180 handover missions, the success rate in vision-based object handover was 79\% (Table II). A likely reason is that the vision-based algorithm is not able to distinguish between pull and push actions. 

\subsection{Fusion: Torque and Vision Data}

\subsubsection{Fusion Algorithm}\label{sec:fusion-function}
The ROS finite state machine (FSM) was used to implement a fusion algorithm to fuse our CNN-based action classifier and SSD fingertips detector. However, we started with implementing a simple "AND" function in our finite state machine package. Fig. \ref{fusion-diagram} shows the fusion diagram. Once the human receiver attempts to receive the object from the robot, the CNN network outputs the related action class (E.g. pulling, pushing, pumping, pulling up, holding, no action). In parallel, our SSD inference package returns the number of detected fingertips on the object and their position with respect to the camera frame. Once all the detected fingers are fallen between the front and back planes of the object, as shown in Fig. \ref{inference_image} (bottom inference image), then the RELEASE decision will be sent to the FSM package, and it gets fused with another command coming from the CNN action classifier using a simple "AND" function. The output of this process is a final RELEASE decision. 
\subsubsection{Experiments}We performed our experiments by using the same number of volunteers for the object handover tasks. Each volunteer was asked to perform six handover experiments. We observed that fusing vision data can significantly improve the accuracy in the final decision made to either release or not release the object. For instance, when the CNN action classifier returned a false positive, the SSD fingertip-detector was able to correct for that and to output the right RELEASE decision. Conversely, when the vision-based algorithm was unable to distinguish between pull and push, the CNN action classifier was able to correct for that and to output the right RELEASE decision. For 180 handover missions (i.e., the total number of experiments done by all human subjects), our success rate was 98\%. Clearly, our fusion algorithm has helped to significantly improve the accuracy in R2H object handover tasks. Our results in Table III show that by adding vision-based finger detection algorithm to the handover system, i.e. fusion algorithm, the success rate improves from 90\% to 98\%. Table IV compares the result of fusion-based handover with some previous works that use success rate as the evaluation metric.

\begin{table}
\centering
\caption{Comparison between our 3 different algorithms in R2H object handover.}
\label{table:3}
\begin{tabular}{|c|c|c|c|}
\hline 
system & Num of trials & Num of successful handovers & s rate \\
\hline
Torque & 180 & 162 & 90\% \\
\hline

 vision & 180 & 114 & 79\% \\ 
\hline
fusion & 180 & 177 & 98\% \\
\hline
\end{tabular}
\end{table}

\begin{table}
\centering
\caption{A comparison with success rates in R2H object handover in previous works.}
\label{table:4}
\begin{tabular}{| m{5em} | m{2cm}| m{2cm} | m{1cm}| }
\hline 
& Number of trials & Number of successful handovers & success rate \\
\hline
 Our method & 180 & 177 & 98\% \\ 
\hline
\cite{choi2009hand} & 144 & 126 & 88\% \\
\hline
 \cite{shi2013model}& 400 & 160 & 40\% \\
\hline
{\cite{grigore2013joint}} & 400 & 160 & 75.6\% \\
\hline
\cite{koene2014experimental}& 44 & N/A & 94\% \\
\hline
\cite{prada2014implementation}& 525 & 495 & 94\% \\
\hline
\end{tabular}
\end{table}

\section{CONCLUSIONS AND FUTURE WORKS}

In this work, we presented a real-time R2H object handover algorithm implemented on a 7-DOF Kinova arm using two sensor modalities of torque and vision. We performed the object handover mission in three scenarios: 1: Contact-, 2: Vision-, and 3: contact-vision-fusion-based. Our contact-based object handover relies on torque data only, which are used as input to a CNN network. The vision-based object handover relies only on the RGB images generated by an eye-in-hand camera, which are used as the input to an SSD object detector to detect and localize the human subject's fingertips in real-time. And finally, our fusion-based algorithm was designed based on a ROS Finite State Machine to fuse both contact and vision-based action detection. Our results show significant improvement in action detection when using our proposed fusion algorithm to perform the R2H handover missions. Fusing vision and torque data in handover missions can improve reliability and safety as well as the robustness of the R2H object handover tasks. 
For future work, we are planning to use deep sensor fusion approaches to unify our network model to a single network, which would decrease the model size and the training time. From the vision perspective, we will incorporate depth information obtained via the RGB-D camera to estimate the fingertips' pose with respect to the camera frame. This can help identify if the fingers are touching the object, and where the contact points would be, if they are. This can also help to estimate the approach velocity of the receiver's hand, which can be viewed as an additional indicator of the receiver's intention.


%
%

\end{document}